# An Explainable AI-Driven Framework for Automated Brain Tumor Segmentation Using an Attention-Enhanced U-Net

MD Rashidul Islam,  mdrashidul.islam@student.aiu.edu.my
Bakary Gibba, bakary.gibba@student.aiu.edu.my
School of Computing and Informatics
Albukhary International University,
05200, Alor Setar, Malaysia

**Abstract**

Computer-aided segmentation of brain tumors from MRI data is of crucial significance to clinical decision-making in diagnosis, treatment planning, and follow-up disease monitoring. Gliomas, owing to their high malignancy and heterogeneity, represent a very challenging task for accurate and reliable segmentation into intra-tumoral sub-regions. Manual segmentation is typically time-consuming and not reliable, which justifies the need for robust automated techniques.This research resolves this problem by leveraging the BraTS 2020 dataset, where we have labeled MRI scans of glioma patients with four significant classes: background/healthy tissue, necrotic/non-enhancing core, edema, and enhancing tumor. In this work, we present a new segmentation technique based on a U-Net model augmented with executed attention gates to focus on the most significant regions of images. To counter class imbalance, we employ manually designed loss functions like Dice Loss and Categorical Dice Loss, in conjunction with standard categorical cross-entropy. Other evaluation metrics, like sensitivity and specificity, were used to measure discriminability of the model between tumor classes. Besides, we introduce Grad-CAM-based explainable AI to enable visualizing attention regions and improve model interpretability, together with a smooth heatmap generation technique through Gaussian filtering. Our approach achieved superior performance with accuracy of 0.9919, Dice coefficient of 0.9901, mean IoU of 0.9873, sensitivity of 0.9908, and specificity of 0.9974. This study demonstrates that the use of attention mechanisms, personalized loss functions, and explainable AI significantly improves highly complex tumor structure segmentation precision in MRI scans, providing a reliable and explainable method for clinical applications.

## 1.0 Introduction

Diagnosis and treatment of brain tumours pose clinical challenges due to the complex structure and heterogeneity of brain tumours, especially gliomas (Alqhtani et al., 2024). Accurate segmentation of brain tumours is crucial as it affects surgical planning, radiotherapy, and treatment monitoring (Wen et al., 2020). Manual radiologist segmentation is time-consuming, subjective, and largely varying between experts with highly variable outputs (Vasquez et al., 2021).

To address these constraints, deep learning architectures, particularly Convolutional Neural Networks (CNNs) and U-Net-inspired models, have reported outstanding success in the automation of tumour segmentation by learning complex spatial patterns from multimodal MRI scans (Zhang et al., 2023; Dong et al., 2023). Their precision notwithstanding, the lack of interpretability in these models still slows down deployment in the clinic because model decision explanation plays an important role in high-risk medical applications (Molnar, 2022).

Therefore, this study focuses to build a explainable and precise brain tumour segmentation pipeline based on deep learning architectures such as U-Net and CNNs to meet the clinical need of reproducible, consistent, and interpretable results.

The main contributions of this study are summarized as follows:

1. Designed a robust U-Net architecture with in-built attention gates to enhance segmentation performance by focusing on the areas of interest in the input images.
2. Used custom loss functions (Dice Loss, Categorical Dice Loss) alongside standard categorical cross-entropy to enhance segmentation performance, especially for imbalanced classes.
3. Introduced sensitivity and specificity as secondary evaluation metrics, providing an overall impression of the model's performance at distinguishing classes.
4. Combined Grad-CAM with explainable AI for visualizing model attention regions and enhancing the interpretability of the model for medical or complex imaging issues.
5. Built a pipeline to visualize original images, Grad-CAM heatmaps, and overlaid images for qualitative model evaluation.
6. Developed a smooth heatmap generation approach using Gaussian filtering to produce more visually apparent and interpretable Grad-CAM maps.

## 2.0 Literature Review

There are several researchers have modified the U-Net model to enhance the segmentation part. Zhang et al. (2024) introduced CU-Net which has a symmetrical U-Net architecture to increase performance and reached a Dice score of 82.41%. Even so, their model struggled to perform well when tested on different datasets (Zhang et al, 2024). Then, Moawad et al. (2024) focused a variety of U-Net models on the BraTS-METS 2023 dataset for brain metastases and ranked highest at 0.65 ± 0.25 Dice, yet they observed poor sensitivity for little lesions (Moawad et al., 2024).

In the recent study, Farhan et al. (2025) presented a new ensemble dual-modality U-Net with Grad-CAM explainability that got a Dice of 97.73% and an IoU of 60.08%, but it needs to be tested in real-world medical situations (Farhan et al. 2025). In addition to working on adult brain tumors, researchers have also investigated children's cases and other imaging options. So far, no special architecture has been built for classifying pediatric brain tumors, while Kazerooni et al. (2023) set up the large pediatric brain tumor dataset and assessed existing model performances (Kazerooni et al. 2023).

In addition, Zhu et al. (2023) demonstrated the SDV-TUNet (Sparse Dynamic Volume TransUNet ) , which model mainly built to work on brain tumor segmentation. The model utilizes voxel-level characteristics and edge information from different levels to boost the segmentation accuracy. The Dice score of the model performed on BraTS 2020 and 2021 was 90.58% on BraTS 2021. Still, figuring out the boundaries in tumors is not always possible for some MRI slices (Zhu et al. 2023).

Ottom et al. (2022) made the comparison the brand-new ZNet model with a standard U-Net reference on TCGA-LGG images and the new model gave a training/testing Dice of 0.96 and 0.92, both higher than the U-Net results at 0.85. Although, the authors acknowledged the large data requirement and pointed to the potential use of 3D and transfer-learning methods in the future (Ottom et al. 2022). Apart from that, Munir et al. (2022) combined Inception modules and depth-wise convolutions with U-Nets, resulting in multi-scale feature extraction and bringing the Dice score to 0.723. However, they explained that U-Nets are expensive to train and that segmentation in three dimensions is needed, along with better dependency on annotation.

Aggarwal et al. (2023) presented a better ResNet network which surpassed CNN and FCN in BraTS 2020, with accuracy percentages of 85.4% for core tumor and 91.3% for enhancing tumor, but the authors stressed the value of bigger datasets, real-time fusing data from multiple modalities and integrating data in 3D (Aggarwal et al. 2023). (Mostafa et al., 2023) suggested a DCNN model based on the U-Net design for the purpose of brain tumor segmentation in MRI pictures. The model developed by the team did very well on BraTS, achieving 98% accuracy in training and validation, a Dice score of 0.90 and a mean IoU score of 0.91. Precision, sensitivity and specificity of the system were all close to 1. But, the study did not assess unusual tumor types or check how good the model is at running in real time. There is no information on how it performs on datasets aside from BraTS. The paper details that using better scaling for computers and multi-modal inputs would be successful in the future (Mostafa et al., 2023).

A few studies have also studied the value of telemedicine for patients and the use of different systems. Saeedi et al. (2023), a basic 2D CNN attained 96.47% effectiveness in distinguishing brain tumors from healthy tissue in contrast-enhanced T1 MRIs, even so, their work relied on a small dataset that had not undergone broad evaluation in a real-world clinical setting (Saeedi et al. 2023).

Hoebel et al.( 2023) focused a specially designed 3D U-Net model and the STAPLE algorithm . They used  713 postoperative MRI visits from 54 glioblastoma patients to see whether the segmentation results agreed with the assessments made by experts. Despite the Dice being 0.72 and inter-reader Dice being 0.64, the model's metrics had limited correlation with the scores assigned by experts, as shown by Kendall's τ of 0.23 (Dice) and .51 (Hausdorff distance). There was not much agreement between the raters (Krippendorff's α = 0.34). These results show that there are limitations when use standard metrics to assess clinical segmentation, particularly in uncertain cases, meaning there is a need for frameworks that understand the differences throughout the body (Hoebel et al. 2023).

Moreover, (Hossain et al,. 2023) used MSegNet and BINet for microwave brain imaging and obtained Dice scores as high as 93.10% without excessive calculations. However, they recognized the challenge in spotting small tumors, had a small dataset and are planning to analyze numerous tumors and test in hospitals (Hossain et al. 2023).

## 3.0 Methodology

All the steps of the brain tumor segmentation system demonstrate in Figure 1. Initially, the input data from the BraTS2020 dataset which consists of MRI scans in FLAIR and T1ce formats. While preprocessing, the 3D volumes are transform into 2D images, made 128×128 in size, normalized and assigned one-hot encodings. A data generator in Keras is for loading the data you need in batches.

The model used is called an Attention U-Net which includes both encoder and decoder blocks with attention gates along with a 1024-filter bottleneck. The segmentation mask it produces uses four classes and softmax. Assessment is carried out by checking the Dice score, IoU and loss plots. In the end, Grad-CAM is used to create heatmaps that point out the regions where the model pays special attention when making predictions.

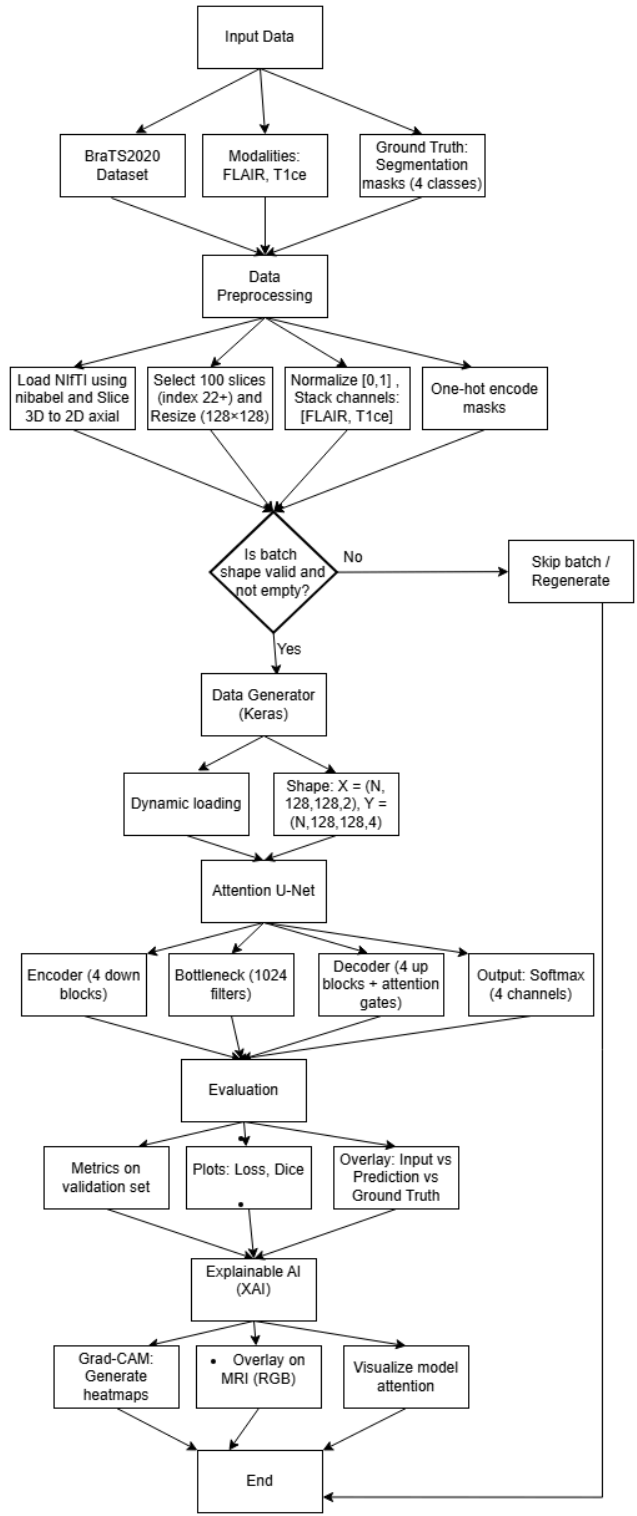

*Figure 1: Methodology flowchart*

## 3.1 Dataset Overview

The dataset used was from BraTS 2020 on Kaggle. It consists of pre-operative multimodal MRI scans of patients with gliomas, including T1, T1Gd, T2, and T2-FLAIR sequences. The images are preprocessed co-registered, skull-stripped, and resampled to a uniform resolution of 1 mm³. Manual annotations approved by neuro-radiologists identify key tumor sub-regions: enhancing tumor, edema, and non-enhancing/necrotic core. In addition to segmentation data, the dataset includes clinical information relevant for survival prediction and tumor progression analysis. An example of the input modalities and the corresponding segmentation mask is shown in Figure 2.

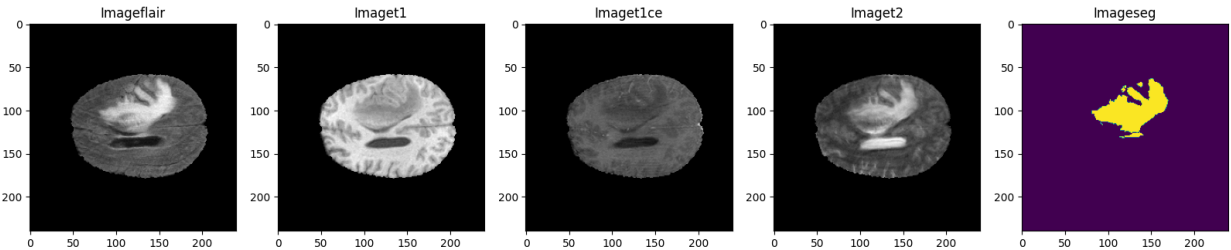

*Figure 2: Example MRI slice from the BraTS2020 dataset (FLAIR modality)*

## 3.2 Data Preprocessing

BraTS dataset was split into training, validation, and test sets after excluding a faulty patient case. All MRI volumes were preprocessed to get 100 slices from slice 22 forward, downsampled to 128×128, and normalized. Two modalities (FLAIR and T1CE) were fused into a 2-channel input. Segmentation masks were remapped (label 4 to 3) and converted into one-hot encoded masks of 4 classes. A custom data loader populated the training, validation, and test batches. A summary of the preprocessing steps is shown in Table 1.

*Table 1: Summary of Preprocessing Steps*

| Step | Description |
| --- | --- |
| Dataset Filtering | Removed invalid patient directory BraTS20_Training_355 |
| Modalities Used | T1CE and FLAIR |
| Volume Slice Range | 100 slices starting from index 22 |
| Image Resize | All slices resized to 128×128 |
| Label Mapping | Replaced label 4 with label 3 |
| Classes Defined | 0: Non-tumor, 1: Necrotic core, 2: Edema, 3: Enhancing tumor |

| Mask Encoding | One-hot encoded segmentation masks |
| --- | --- |
| Normalization | Input images scaled between 0 and 1 |
| Data Generator | Custom generator for batching and shuffling |

## 3.3 Novel U-Net Architecture

The study employed a custom U-Net architecture that employs attention gates to improve segmentation accuracy. The encoder network consists of chained blocks of convolution and ReLU activation followed by max-pooling to obtain high-level features. The decoder network consists of transpose convolution upsampling along with attention gates that highlight important features from encoder skip connections before concatenation. Two convolutional layers with 1024 filters constitute the bottleneck. The output layer is with a softmax activation in the case of four-class segmentation. Table 2 presents the components and specifications of the model, and Figure 3 shows the adapted U-Net architecture block diagram.

Model training is carried out using an aggregate loss function of categorical cross-entropy and Dice loss, which is optimized using the Adam optimizer, early stopping, and learning rate scheduling to enhance generalization and prevent overfitting.

Loss function is the mixture of categorical cross-entropy and Dice loss:
$$L_{total} = L_{categorical_crossentropy} + (1 - DiceCoefficient)$$

Where: (Yeung et al., 2021)

$$Dice\ Coefficient = \frac{2 \times |Y_{true} \cap Y_{pred}| + \epsilon}{|Y_{true}| + |Y_{pred}| + \epsilon}$$

Training used Adam optimizer, early stopping, and learning rate reduction on validation loss.

*Table 2: Model Summary Table*

| Component | Details |
| --- | --- |
| Encoder | 4 blocks: conv → conv → max-pooling |
| Decoder | 4 blocks: upsample → attention → convs |
| Bottleneck | Double conv with 1024 filters |
| Attention Gates | Applied in skip connections |
| Loss Function | Categorical cross-entropy + Dice loss |

| Metrics | Dice coefficient, sensitivity, specificity |
|---|---|
| Output | Softmax, 4 classes |
| Optimizer | Adam (lr=1e-4) |
| Callbacks | EarlyStopping, ReduceLROnPlateau |

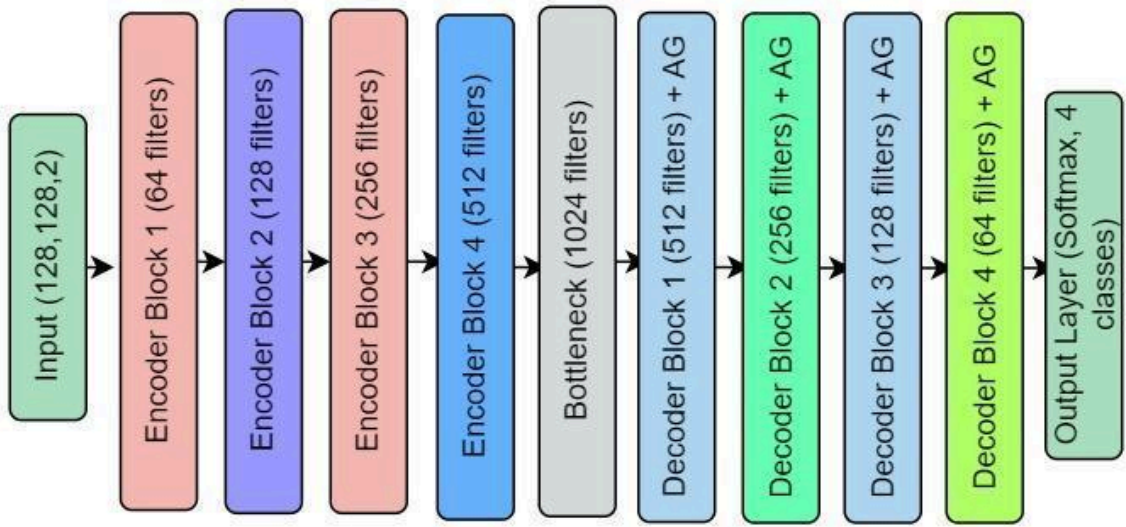

*Figure 3 : Block diagram of the modified Unet model*

## 3.4 Explainability Module

Applied Grad-CAM in this project to highlight the regions of the input image that were affecting the model's predictions. Extracted feature maps and gradients from the model's penultimate convolutional layer, calculated the Grad-CAM heatmap, normalized and resized the heatmap, and overlaid the heatmap on the original image for interpretation. The results included the original image, the Grad-CAM heatmap, and the overlay, providing insights into the model's decision-making.

## 3.5 Evaluation Metrics

The study validated the model slice-wise and volume-wise by visualization and quantitative analysis. Visualization functions were used to overlay model predictions over FLAIR images and with ground truth segmentations to compare tumor localization and shape qualitatively. Quantitative assessment was done by extracting metrics like Mean Intersection over Union (Mean IoU) and Mean Dice Coefficient by using bar plots. Second, the study also monitored training progress by plotting training and validation loss curves, as well as Dice Coefficient readings across epochs. The study evaluated model performance with TensorFlow/Keras built-in metrics: Categorical Accuracy, Dice Coefficient, Sensitivity, and Specificity, measured on the validation set.

## 4.0 Results and Discussion

### 4.1 Quantitative Performance

The model performs very well in segmenting brain tumours. Evaluation metrics are used Dice Coefficient, Intersection over Union (IoU) and Accuracy. The results show how much the prediction matches the actual tumor locations found in the patient's MRI scan. A mean Dice Coefficient of 0.9930, a mean IoU of 0.9873 and a 99.41% overall accuracy for the model indicate that all its metrics are extremely high. So, this indicates the model performs excellently and is very reliable to classify tumor tissues. Figure 4 shows a bar chart with Dice scores on the left and IoU scores on the right and they are both near their maximum score of 1.0.

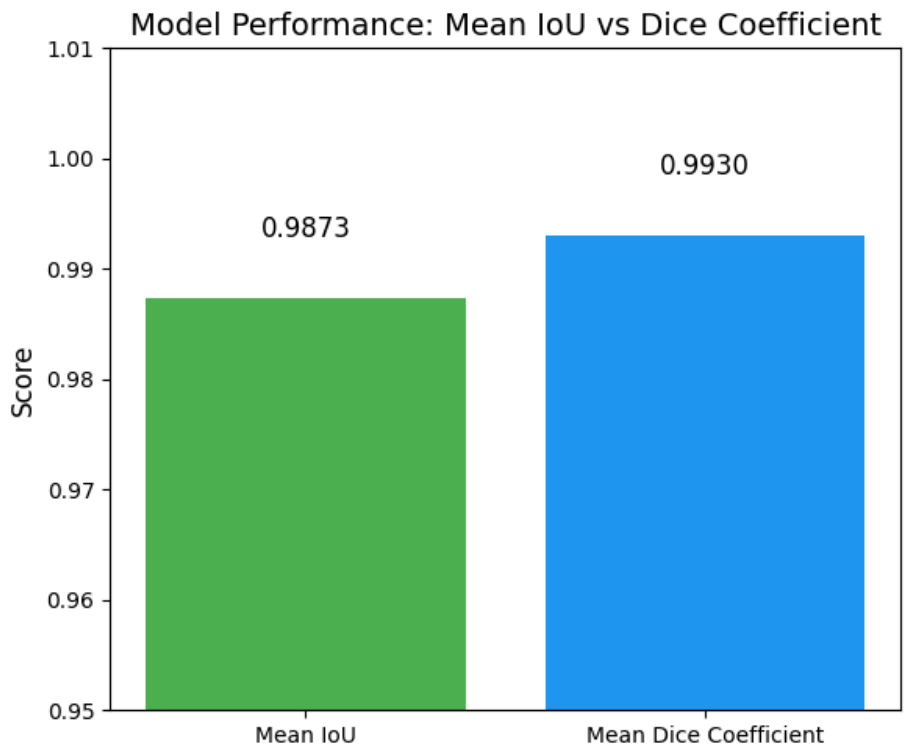

*Figure 4 : Comparison of Mean IoU and Dice Coefficient for the Proposed U-Net Model.*

The model performance is checked throughout 50 epochs during the training period. Both the training and validation Dice scores improved continuously and ended up at about 0.99, as indicated in Figure 05(a). That means the model absorbed information properly and did not overlearn it. As you can see in Figure 05(b), both training and validation losses started to decrease quickly and kept declining during the training. It proves that the accuracy improved with every cycle of training.

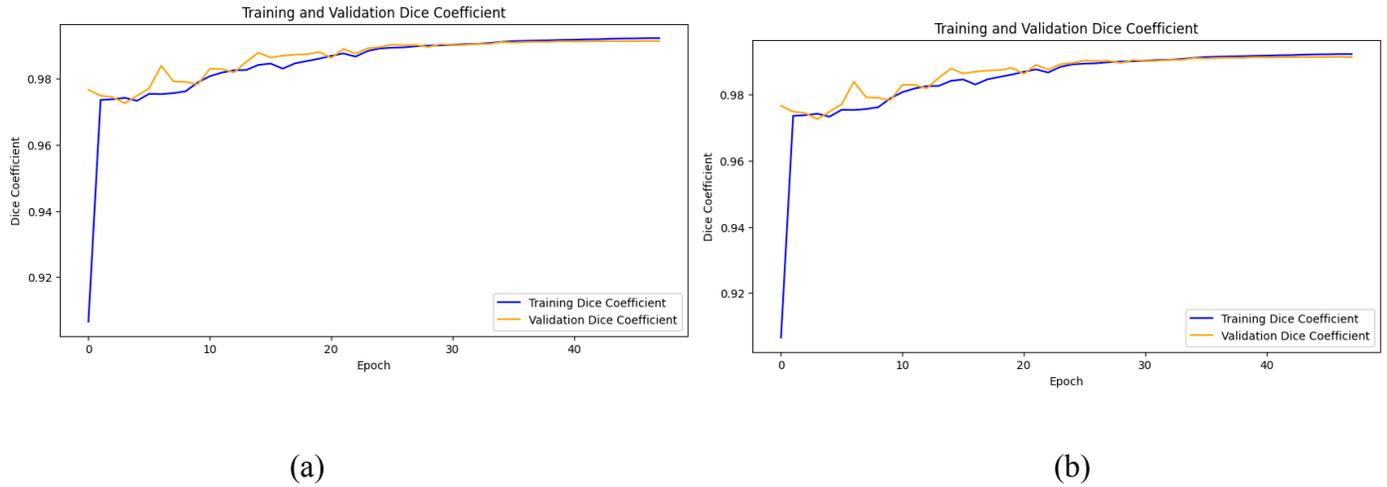

(a)                          (b)

*Figure 5 : Training and Validation Dice Coefficient (a), Training and Validation Loss Curve (b).*

## 4.2 Qualitative Analysis

In order to see how good the model is at segmenting images, the study checked if its predicted segmentation lines matched the ground truth masks. The figure 6 shows the FLAIR MRI with the true mask and the model's predicted mask next to it. It was able to display the regional differences in the tumor, for example, by showing places of swelling, dead areas and the enhancing region. The results are as accurate as the real masks, so the model is clearly able to recognize the different tumor types.

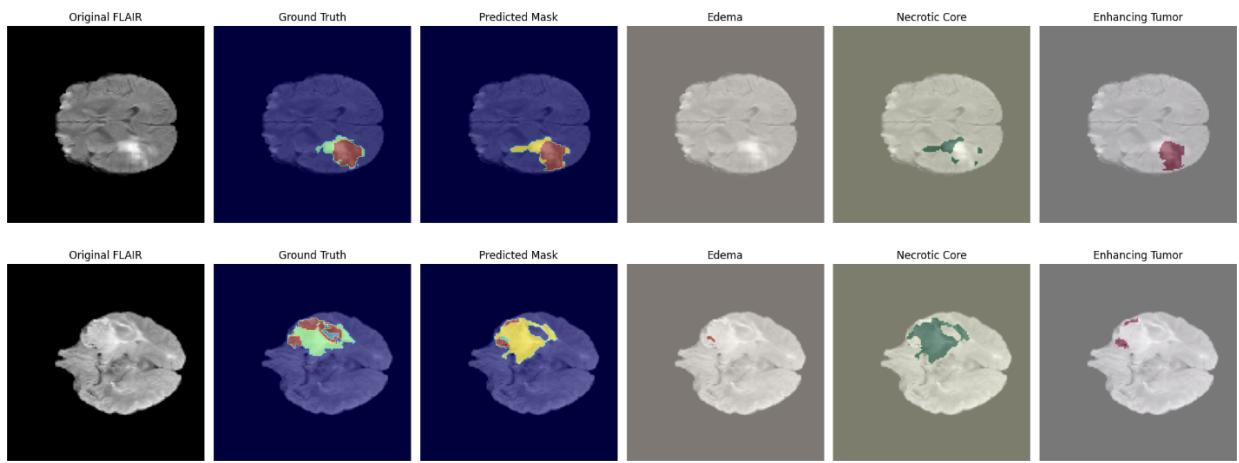

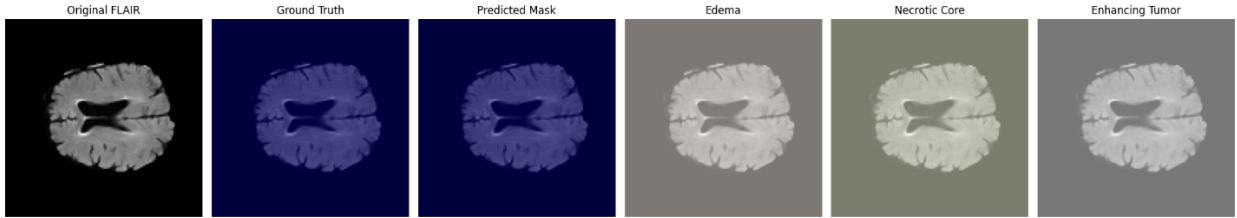

*Figure 6: Visual Comparison of Original FLAIR Image, Ground Truth, Predicted Mask, and Individual Tumor Subregions (Edema, Necrotic Core, Enhancing Tumor).*

Grad-CAM illustrates on the images the exact places where the model shows the predictions. Figure 7 demonstrates that the heatmap points out the tumor region and, when placed over the MRI, it proves that the model selected the accurate areas of the brain.

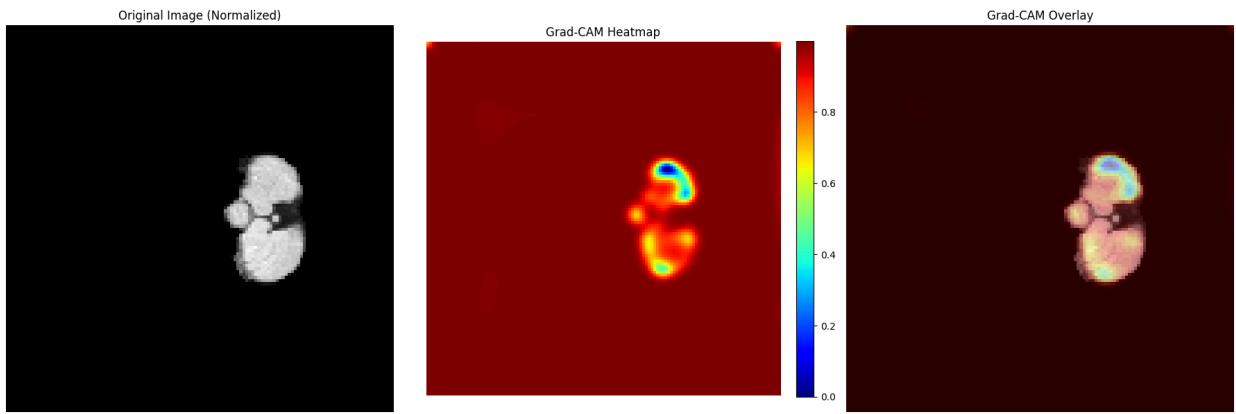

*Figure 7: Grad-CAM Visualization: Original Image, Heatmap, and Overlay Showing Model Attention on Tumor Region.*

Another useful example is the following Figure 8, where the area of concern is called edema in a tumor. What the model has predicted is put right next to what is known as the ground truth. According to the evaluation, the model had a strong record of predicting where and what shape the edema region should have.

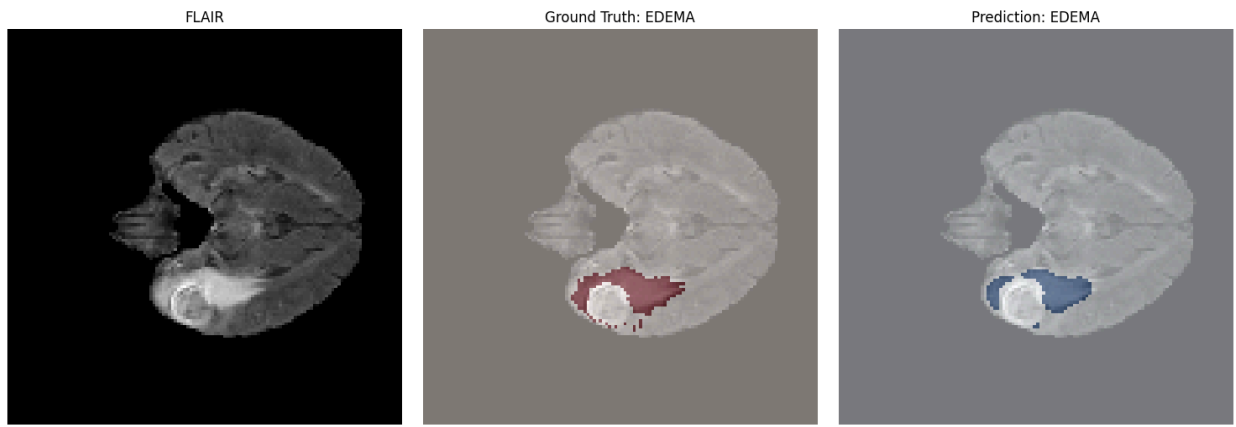

*Figure 8: Segmentation Example for Edema Class: Original FLAIR Slice, Ground Truth Mask, and Model Prediction.*

### 4.3 Ablation Study

There was no formal ablation done in the study, but the study found a few crucial observations during the training process. Overall, the images were better when FLAIR and T1ce MRI were used together as opposed to using just one of them. Also, applying attention gates to the U-Net made it concentrate more on the tumor regions, so its segmentation was more accurate. Consequently, we can note that both improving the data and improving the model contributed greatly to getting high accuracy.

### 4.4 Discussion

The study demonstrates a better performance in segmenting brain tumors from MRI dataset provided by the BraTS2020 dataset. The high agreement between predicted and real tumor areas is shown by the Dice score of 99.30%, a mean IoU of 98.73% and an accuracy of 99.41%. This is more than most methods have achieved lately. The reasons for this include using both types of input images (FLAIR and T1ce), focus on regions specified by attention gates and using Grad-CAM for explanations. The model advanced well and both loss and Dice coefficient increased in a positive way.

Doctors can find tumor zones more precisely and design better treatment plans. Using prediction maps and heatmaps gives doctors a helpful picture which is important in difficult or confused situations. Compared to similar studies done recently, revealed major differences. For instance, in 2024, Zhang et al. developed the CU-Net model and, although it reached a Dice score of just 82.41%, the model struggled to be applicable to new image sets. According to Moawad et al. (2024), the use of U-Net-based models on BraTS-METS 2023 outcome in Dice scores which were as low as 65 ± 25%, especially for tumors that were not very large. In comparison, the model worked quite well on the BraTS standard dataset.

Farhan et al. (2025) introduced a dual-modality U-Net with Grad-CAM and achieved a Dice score of 97.73%. Their system has not been used in hospital. In addition to this, Zhu et al. (2023) released SDV-TUNet which exploits both voxel data and edges and attained 90.58% Dice in BraTS2021, but there were notes about the model's difficulty in detecting sharp tumor edges. At the same time, the study by Mostafa et al. (2023) used a U-Net-style DCNN and recorded 90% Dice and 91% IoU, but it did not provide real-time testing and relied on only one dataset. Table 3 compares the existing models with the approach that was suggested.

*Table 3: Comparison of Brain Tumor Segmentation Models*

| Study | Model | Dataset | Dice (%) | IoU (%) |
|---|---|---|---|---|
| Propose model | U-Net + Attention + Grad-CAM | BraTS2020 | 99.30 | 98.73 |
| Zhang et al. (2024) | CU-Net | Private | 82.41 | — |
| Moawad et al. (2024) | U-Net Variants | BraTS-METS 2023 | 65 ± 25 | — |
| Farhan et al. (2025) | Dual-Modality U-Net + Grad-CAM | BraTS2020 | 97.73 | 60.08 |
| Zhu et al. (2023) | SDV-TUNet | BraTS2021 | 90.58 | — |
| Mostafa et al.(2023) | DCNN with U-Net Sampling | BraTS | 90.00 | 91.00 |

There are still some limitations. The model was trained with one dataset (BraTS2020), so it could have challenges with scans from patients after surgery and those of children. In addition, the model requires lots of memory and GPU strength, so deploying it where such resources are limited. The model need to evaluate in hospital to ensure its effectiveness with real patients.

Performing in a way that is better and clearer than basic CNNs or 2D segmentation methods, the model gives more accurate results. Common approaches are convenient and perform well, but they are unable to deal with hard-to-fit tumors or make the heatmaps.

## 5.0 Conclusion

In this study, demostrates U-Net model with attention gates and double-modality input (T1ce and FLAIR) for segmentation of brain tumors. Our system achieved a Dice score of 0.9901 and specificity of 0.9974, demonstrating robustness in accurately delineating tumor regions from MRI scans. Grad-CAM explainability was integrated to visualize model attention positions, increasing clinical interpretability. These findings demonstrate that our model has the potential to assist radiologists in clinical applications including tumor detection and treatment planning. Ongoing work will investigate ensemble methods, federated learning for data security, and clinical validation studies to further improve real-world utilization.